\documentclass[letterpaper, 10 pt, conference]{ieeeconf}  % Comment this line out if you need a4paper

\IEEEoverridecommandlockouts                              % This command is only needed if 
\usepackage{graphics} % for pdf, bitmapped graphics files
\usepackage{epsfig} % for postscript graphics files
\usepackage{mathptmx} % assumes new font selection scheme installed
\usepackage{times} % assumes new font selection scheme installed
\usepackage{amsmath} % assumes amsmath package installed
\usepackage{amssymb}  % assumes amsmath package installed
\usepackage{enumerate}
\usepackage{bm}
\usepackage{graphicx}
\usepackage{epstopdf}
\usepackage{array}
\title{\Large \bf Smart Explorer: Recognizing Objects in Dense Clutter via Interactive Exploration
}

\author{Zhenyu Wu$^{1}$, Ziwei Wang$^{2}$, Zibu Wei$^{2}$, Yi Wei$^{2}$ and Haibin Yan$^{*1}$% <-this % stops a space
\thanks{*Corresponding author. The first two authors contribute equally}% <-this % stops a space
\thanks{Code: https://github.com/Gary3410/Smart-Explorer}
\thanks{$^{1}$Zhenyu Wu and Haibin Yan are with the School of Automation, Beijing University of Posts and Telecommunications, Beijing, 100876, China. {\tt\small \{wuzhenyu, eyanhaibin\}@bupt.edu.cn}}        
\thanks{$^{2}$Ziwei Wang, Zibu Wei and Yi Wei are with the Department of Automation, Tsinghua University, and Beijing National Research Center for Information Science and Technology (BNRist), Beijing, 100084, China.  {\tt\small \{wang-zw18, weizb18, y-wei19\} @mails.tsinghua.edu.cn.}}
}

\begin{document}

\maketitle
\thispagestyle{empty}
\pagestyle{empty}

%%%%%%%%%%%%%%%%%%%%%%%%%%%%%%%%%%%%%%%%%%%%%%%%%%%%%%%%%%%%%%%%%%%%%%%%%%%%%%%%
\begin{abstract}
Recognizing objects in dense clutter accurately plays an important role to a wide variety of robotic manipulation tasks including grasping, packing, rearranging and many others. 
However, conventional visual recognition models usually miss objects because of the significant occlusion among instances and causes incorrect prediction due to the visual ambiguity with the high object crowdedness. 
In this paper, we propose an interactive exploration framework called Smart Explorer for recognizing all objects in dense clutters. Our Smart Explorer physically interacts with the clutter to maximize the recognition performance while minimize the number of motions, where the false positives and negatives can be alleviated effectively with the optimal accuracy-efficiency trade-offs. 
Specifically, we first collect the multi-view RGB-D images of the clutter and reconstruct the corresponding point cloud. By aggregating the instance segmentation of RGB images across views, we acquire the instance-wise point cloud partition of the clutter through which the existed classes and the number of objects for each class are predicted. The pushing actions for effective physical interaction are generated to sizably reduce the recognition uncertainty that consists of the instance segmentation entropy and multi-view object disagreement. Therefore, the optimal accuracy-efficiency trade-off of object recognition in dense clutter is achieved via iterative instance prediction and physical interaction. Extensive experiments demonstrate that our Smart Explorer acquires promising recognition accuracy with only a few actions, which also outperforms the random pushing by a large margin.
\end{abstract}

%%%%%%%%%%%%%%%%%%%%%%%%%%%%%%%%%%%%%%%%%%%%%%%%%%%%%%%%%%%%%%%%%%%%%%%%%%%%%%%%
\section{Introduction}
Robotic manipulation tasks have been considered to be challenging due to the complexity caused by the severe occlusion and high crowdedness in dense clutter, which usually appears in the practical working environments such as warehouses. Accurately recognizing all objects in clutter is general requirements for a wide variety of robotic manipulation tasks including grasping \cite{mousavian20196}, \cite{sundermeyer2021contact}, \cite{2022arXiv220711941L}, packing \cite{wang2020robot}, \cite{wang2021dense}, \cite{mitash2020task}, rearranging \cite{cheong2020relocate}, \cite{wang2021uniform}, \cite{danielczuk2021object} and many others. For example, the packing robot is able to present suitable packing orders and spatial locations in boxes only when perceiving accurate information of existed classes and corresponding object numbers for each class \cite{wang2021dense}. Therefore, it is desirable to design the visual perception model that is capable of accurately recognizing all objects in clutter.

In order to strengthen the recognition performance for objects in dense clutter, visual detection and segmentation models specifically designed for clutter scenes were presented \cite{xiang2021learning}, \cite{xie2020best}, where the discriminality of features for overlapped regions was enhanced for retrieving objects with significant occlusion. Because of invisible objects that are completely blocked, the performance is still far from expected for deployment. The active exploration frameworks \cite{ye2021efficient}, \cite{novkovic2020object}, \cite{bejjani2021occlusion} were proposed to search target objects by a series of actions such as pushing and grasping, where the learned planner adaptively selects the optimal spatial location for the actions according to observation of current clutter scene. Nevertheless, the existing active exploration methods can only search the given target object, which is not applicable for recognizing all objects in clutter since the target is not determined until the environment is fully explored. Moreover, the complicated planner in conventional active exploration methods results in heavy computational cost and sizably degrades the efficiency of robot manipulation.

\begin{figure}[t]
	\centering
	\includegraphics[height=3.7cm, width=8.7cm]{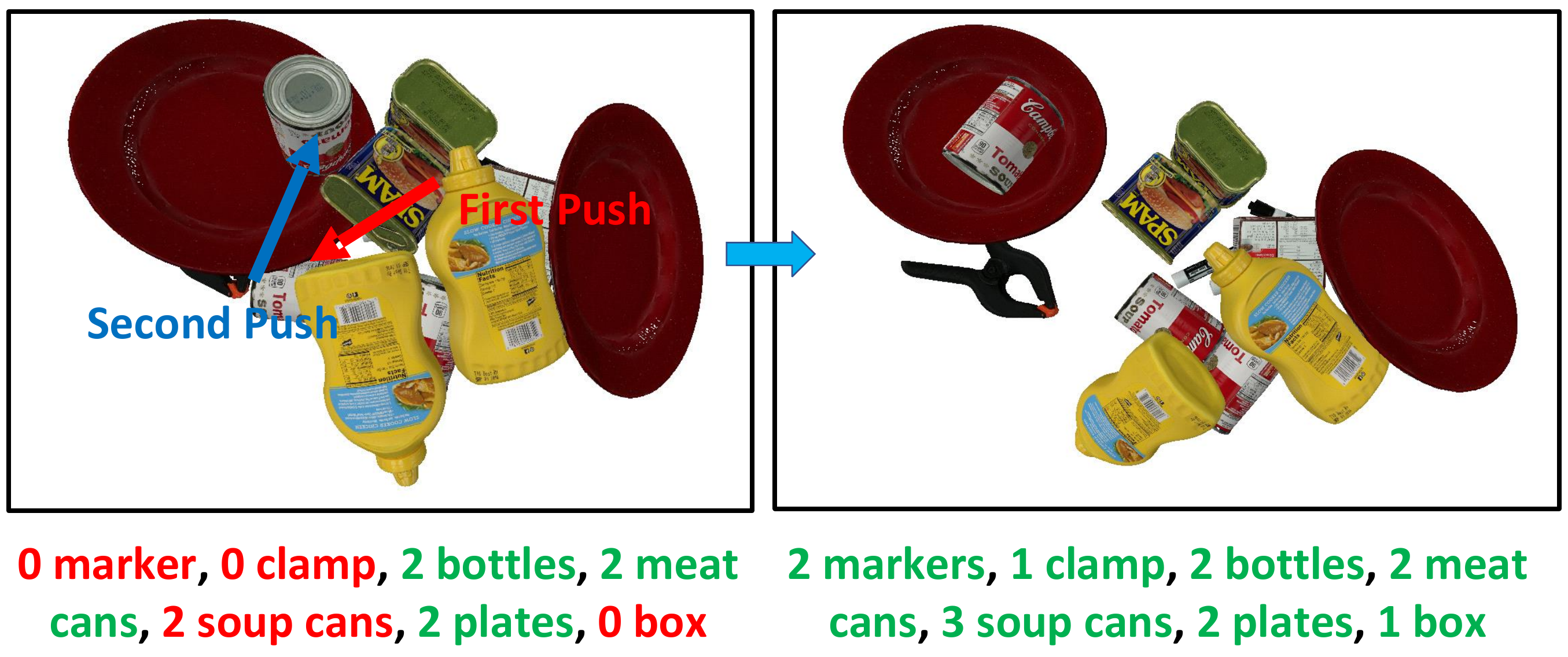}
	\caption{An example of interactive exploration for object recognition in dense clutter. Object recognition in clutter aims to predict the existed classes and the number of objects for each class, which are demonstrated by the text below the images. Green text represents the correct prediction including both the class and the number of objects, while the red text means the incorrect one. By iteratively pushing the clutter, the object recognition is significantly enhanced by discovering the invisible objects with visual ambiguity elimination.}
	\vspace{-0.5cm}    
	\label{comparison}
\end{figure}

In this paper, we propose the Smart Explorer framework to interactively explore the clutter for recognizing all objects accurately. Our Smart Explorer generates the pushing actions for the clutter with the goal of maximizing the recognition performance while minimizing the number of motions, so that the false positives and negatives of recognizing all objects in clutter are effectively alleviated with the optimal accuracy-efficiency trade-offs. More specifically, we collect the multi-view RGB-D images of the clutter and reconstruct the point cloud, which are leveraged as the visual input of our Smart Explorer. Then we employ the instance segmentation model for the multi-view RGB images, which is aggregated to yield the instance partition of point cloud for the clutter. Therefore, the existed classes and the number of objects for each class are predicted according to the instance segmentation for the point cloud. Since reducing the uncertainty provides more information for object recognition in clutter, we generate the start point, orientation and distance of the pushing actions based on the recognition uncertainty that consists of instance segmentation entropy and multi-view object disagreement. Moreover, we also evaluate the spatial constraint between pushing actions and their interacted regions to guarantee the validity of the physical interaction, so that the clutter structure can be effectively modified by physical interaction. 
Fig. \ref{comparison} demonstrates an example of our interactive exploration for object recognition in dense clutter. Compared with recognizing objects in dense clutter directly, experimental results on object recognition show that our Smart Explorer enhances the recall and precision rates by $16.13\%$ and $9.47\%$ respectively with only $1.67$ pushing actions on average, which also outperforms the random pushing by a large margin. Our contribution can be summarized as follows:

\begin{enumerate}[(1)]
	\item To the best of our knowledge, we propose the first framework for recognizing all objects in dense clutter based on the multi-view RGB-D cameras, which provides necessary object information for a wide variety of robotic manipulation tasks such as grasping, packing and rearranging.
	
	\item We present the interactive exploration method that actively pushes the clutter to reduce the prediction uncertainty with spatial relation constraint, which significantly enhances the recognition accuracy with the optimal accuracy-efficiency trade-off.
	
	\item We conduct extensive experiments on the task of object recognition in dense clutter, and the results indicate our Smart Explorer can outperform random pushing by a large margin and sizably increase the recognition accuracy with only a few actions.
\end{enumerate}

\section{Related Work}
\textbf{Object recognition in cluttered scenes: }Existing object recognition for cluttered objects methods can be divided into two categories: recognition with RGB-D images and point cloud. For the first regard, many robotic grasping techniques \cite{wada2019joint}, \cite{yang2020deep}, \cite{schwarz2018rgb} were equipped with a visual segmentation module for the planner to select the grasp pose. In order to segment unseen objects in the clutter for more reliable visual perception, Xie \emph{et al.} \cite{xie2020best} first acquired the rough initial mask based on the depth image and then refined the prediction according to the RGB images. They further utilized graph neural networks to mine the relationship among objects for more accurate instance mask refinement \cite{xie2022rice}. Xiang \emph{et al.} \cite{xiang2021learning} extracted pixel-wise feature embedding with metric learning techniques, where mean shift clustering was employed to discovered unseen objects. For object recognition in cluttered scenes based on point cloud, Tombari \emph{et al.} \cite{tombari2010object} deployed 3D detection and description methods to match the correspondence between the template and the scene, followed by Hough vote for evidence accumulation of centroids. Buch \emph{et al.} \cite{buch2017rotational} used geometric constraints to cast full 6 DoF votes with individual correspondence to enhance the robustness to outliers. Tao \emph{et al.} \cite{tao2020pipeline} leveraged the local shape descriptors with clustering to select the correct correspondence, where the cluster index was developed to verify the transformation hypothesis. However, the severe occlusion causes invisible objects that are completely blocked, which significantly degrades the performance of object recognition in clutter scenes. Our method physically interacts with the clutter for occlusion alleviation, where the recognition performance is maximized with least motions to achieve the optimal accuracy-efficiency trade-off.

\textbf{Active exploration in visual perception: }As part of active learning \cite{wang2020deep}, active exploration has been widely studied in visual perception of robotic tasks such as object segmentation \cite{eitel2019self}, \cite{xu2015autoscanning}, \cite{patten2018action}, \cite{singh2021nudgeseg}, mapping \cite{julian2014mutual}, \cite{popovic2017multiresolution}, \cite{bourgault2002information}, \cite{rocha2005cooperative} and target search \cite{ye2021efficient}, \cite{novkovic2020object}, \cite{bejjani2021occlusion}, \cite{danielczuk2019mechanical}, \cite{nam2020fast}, which observes and manipulates the object clutter adaptively to extract discriminative information of the scene for subsequent downstream tasks. For object segmentation, Patten \emph{et al.} \cite{patten2018action} employed probabilistic segmentation framework where the uncertainty was used to guide the robot for scene manipulation, and the scene motion provided additional clue for associating observed parts to objects. Early works for mapping \cite{bourgault2002information}, \cite{rocha2005cooperative}, \cite{julian2014mutual} targets at maximizing the map information gain via evaluating the mutual information. Popovic \emph{et al.} \cite{popovic2017multiresolution} fused the multi-resolution data with Gaussian Process as priors to increase the efficiency of informative path planning for UAVs. Liu \emph{et al.} \cite{liu2021active} utilized the dynamic Gaussian Process Implicit Surface method \cite{williams2006gaussian} to incrementally update the scene map for mobile picking, and the next-best-view was calculated by balancing the object reachability for picking and map information gain for fidelity and coverage. Target search aims at locating and extracting a known object from the clutter. Danielczuk \emph{et al.} \cite{danielczuk2019mechanical} modeled the target search as a Partially Observable Markov Decision Process, where the actions of push, suction and grasp were iteratively performed until extracting the target object. The difference between our task and target search is that we aim at recognizing all objects without specific targets, which is more challenging without the priors acquired from the pre-defined targets.

\begin{figure*}[t]
	\centering
	\includegraphics[height=5.8cm, width=17.5cm]{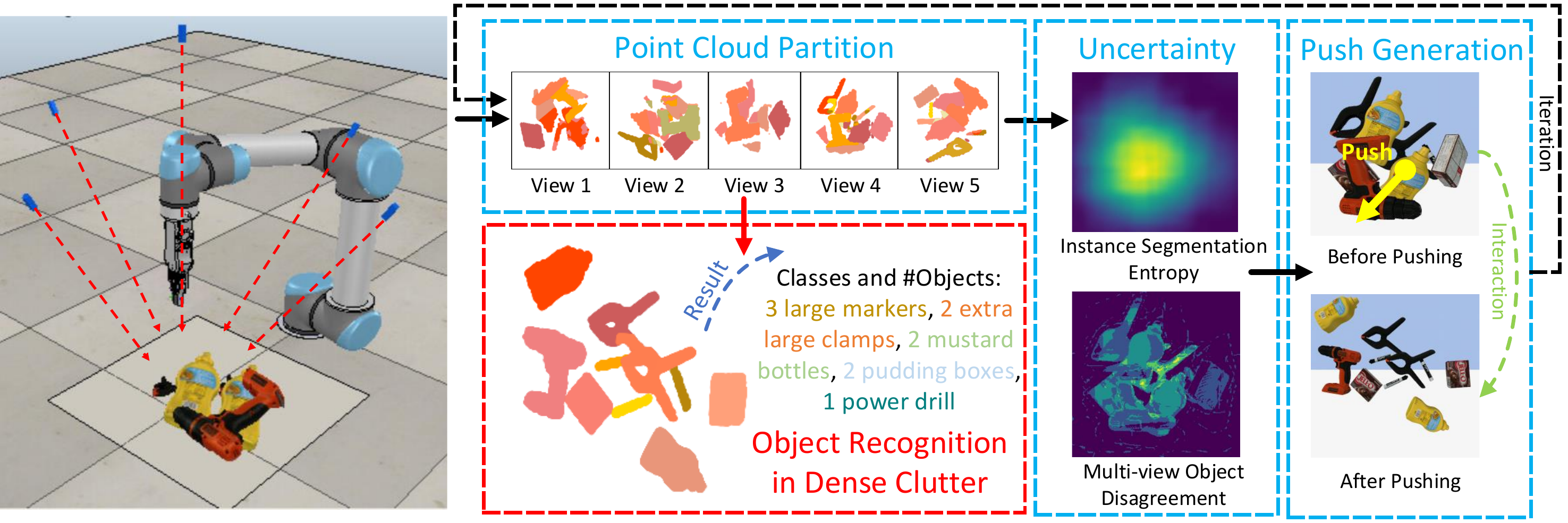}
	\caption{The pipeline of our Smart Explorer. The clutter is observed by one overhead and four side-view cameras, and the point cloud collected across views are partitioned according to the 2D instance segmentation. Moreover, the recognition uncertainty including the instance segmentation entropy and multi-view object disagreement is computed for the clutter in the top-down direction (vertical view), and the effective pushing actions are generated according to recognition uncertainty. By iterative object recognition and physical interaction, the object recognition is achieved via predicting the existed classes and the number of objects for each class based on the instance segmentation of the point cloud.}
	\vspace{-0.5cm}
	\label{pipeline}
\end{figure*}

\section{Recognizing Objects with Interactive Exploration}
In this section, we first briefly introduce the pipeline overview of object recognition in dense clutter, and then detail the pipeline of object recognition including 2D instance segmentation and instance consistency. After that, we formulate the recognition uncertainty based on instance segmentation entropy and multi-view object disagreement. Finally, we present the optimal pushing generation for interactive exploration by recognition uncertainty and spatial constraint evaluation.

\subsection{Pipeline Overview}
The objective of object recognition in clutter is to predict the existed classes and the corresponding number of objects for each category, which provides object information for downstream manipulation tasks such as packing and rearranging. To enhance the recognition performance in the environment with significant occlusion and crowdedness, we enable the agent to physically interact with the clutter so that more informative clues are acquired with active exploration. Meanwhile, the number of actions should be minimized to achieve the optimal accuracy-efficiency trade-off.

Fig. \ref{pipeline} depicts the overall pipeline of our Smart Explorer. The clutter is observed by one overhead and four side-view RGB-D cameras, and the side-view cameras are uniformly placed in a horizontal plane. The point cloud of the clutter is obtained by projecting that converted from the depth image in each view to the world space, which is combined with the multi-view RGB images as the input for recognizing objects in clutter. We leverage the instance segmentation model to obtain the 2D object mask in the multi-view RGB-D images, which assigns object labels to the clutter point cloud projecting inside the mask for each view. The instance segmentation mask for point cloud of each object is generated by merging point cloud partitions across views with instance consistency. Consequently, the existed classes and the number of objects for each class are predicted based on instance segmentation of the clutter point cloud.

We employ pushing as the action primitive in the interactive exploration, which is implemented by the closed gripper that moves along with a straight line in parallel with the tabletop. Since locations with recognition uncertainty composed of instance segmentation entropy and multi-view object disagreement indicates prediction ambiguity, the start point, pushing direction and pushing distance are generated based on the recognition uncertainty with spatial constraint evaluation to provide the informative visual clues with high motion efficiency. By iterative object recognition and physical interaction, the recognition performance for cluttered scenes is significantly enhanced with alleviated false positives and negatives.

\subsection{Object Recognition for Cluttered Scenes}
We leverage the multi-view RGB-D cameras to obtain more informative visual clues with larger fields. The point cloud of the clutter is reconstructed from that collected from each view. Directly utilizing the 3D point cloud segmentation framework for instance partition fails to generalize on objects with shape variation. Therefore, we employ the 2D instance segmentation framework for RGB images across views to assign object labels to clutter point cloud projecting inside the mask. Since the instance masks from different views may represent the same object, the object consistency across views should be verified to avoid false positives and negatives in instance segmentation of clutter point cloud.

\begin{figure}[t]
	\centering
	\includegraphics[height=4.1cm, width=8.7cm]{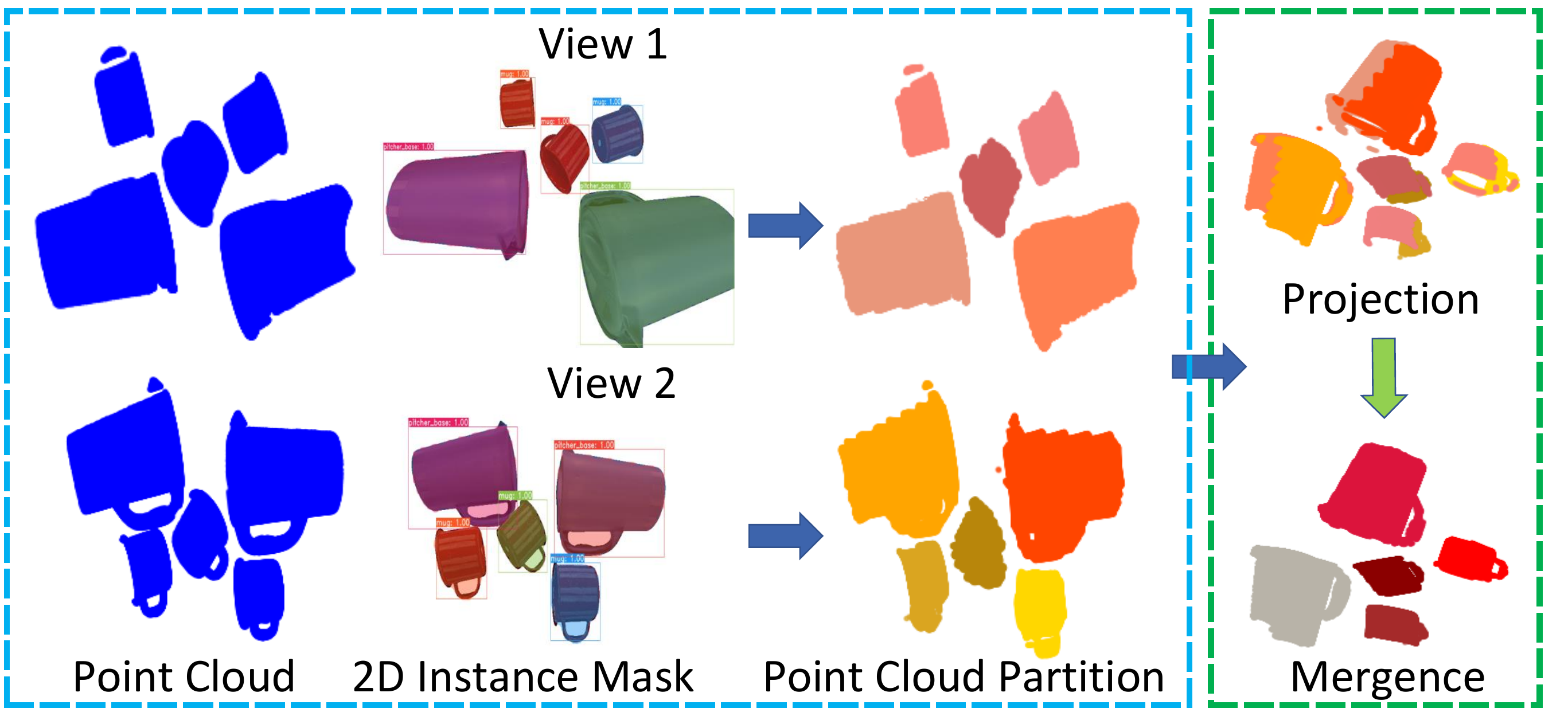}
	\caption{Demonstration of object recognition for cluttered scenes. For each view, the instance segmentation model for RGB images obtains the object masks to assign instance labels to the point cloud. The point cloud partitions across different views are projected to the world space, and those with the same semantic labels and similar spatial occupancy are merged to acquire the instance segmentation masks of the overall point cloud. Therefore, the existed classes and the corresponding number of objects for each category obtained via the instance segmentation of point cloud are utilized as the results of object recognition in clutter.}
	\vspace{-0.5cm}    
	\label{merge}
\end{figure}

We verify the object consistency for different point cloud partitions across views according to the predicted categories and geometric relationship, which can be iteratively merged for those sharing the same semantic labels and similar spatial occupancy in order to acquire the instance segmentation masks of the overall point cloud. 
%Let us denote the class of the $i_{th}$ point of the point cloud in the $k_{th}$ view as $c_{k,i}^{po}$, which is defined based on the corresponding instance segmentation mask category in the RGB images from the $k_{th}$ view:
%\begin{align}
%	c_{k,i}^{po}=\sum_{c=1}^{C}c\cdot I(c_{k,i}^{im}=c)
%\end{align}where $c_{k,i}^{im}$ represents the class of the projected pixel in the RGB images for the $i_{th}$ point of the point cloud in the $k_{th}$ view, and $C$ stands for the number of possible classes of objects in the clutter. Meanwhile, $I(x)$ means the indicator function equal to one for true $x$ and zero otherwise. The overall point cloud collected from each view is partitioned into several parts according to the instance segmentation, and the point cloud partitions in different views is iteratively merged when they share the same semantic labels and spatial occupancy in order to acquire the instance segmentation masks of the overall point cloud. 
Denoting the point cloud of the $i_{th}$ instance in the clutter at the $t_{th}$ merging iteration as $\mathcal{P}_i^{t}$, the instance segmentation mask of the point cloud is updated in the following:
\begin{align}\label{iterative_merge}
	\mathcal{P}_i^{t+1} = \mathcal{P}_i^{t} \cup \{\bm{S}_{m}^{k}|c_m^k=C_i^0, d_{ch}(\bm{S}_{m}^{k}, \mathcal{P}_i^{t})<h\} 
\end{align}where $\bm{S}_{m}^{k}$ means the $k_{th}$ point cloud partition in the $m_{th}$ view.
Meanwhile, $c_m^k$ and $C_i^0$ demonstrate the predicted class of $\bm{S}_{m}^{k}$ and $\mathcal{P}_i^{t}$ respectively. The Chamfer distance between $\bm{x}$ and $\bm{y}$ is denoted as $d_{ch}(\bm{x},\bm{y})$, which measures the spatial difference between the two point sets. The hyperparameter $h$ means the threshold where point sets with Chamfer distance smaller than $h$ are regarded from the same instance. Each point cloud partition across views is regarded as individual instance at the initialization of merging, and the iteration ends until no instance is enlarged by point cloud merging. Finally, the point cloud of the clutter is segmented into multiple instances with predicted labels. Fig. \ref{merge} illustrates the point cloud mergence across views. Therefore, the existed classes and the corresponding number of objects for each category obtained via the instance segmentation of point cloud are utilized as the results of object recognition in clutter.

\subsection{Uncertainty of Object Recognition in Clutter}
High recognition uncertainty indicates ambiguous visual clues in perception, which usually leads to incorrect predictions without sufficient confidence. Reducing recognition uncertainty benefits object recognition in dense clutter by enhancing prediction informativeness. Since we acquire the prediction of object recognition in dense clutter via 2D instance segmentation and object consistency verification, we define the recognition uncertainty based on the instance segmentation entropy and the multi-view object disagreement.

\begin{figure}[t]
	\centering
	\includegraphics[height=9.3cm, width=8.8cm]{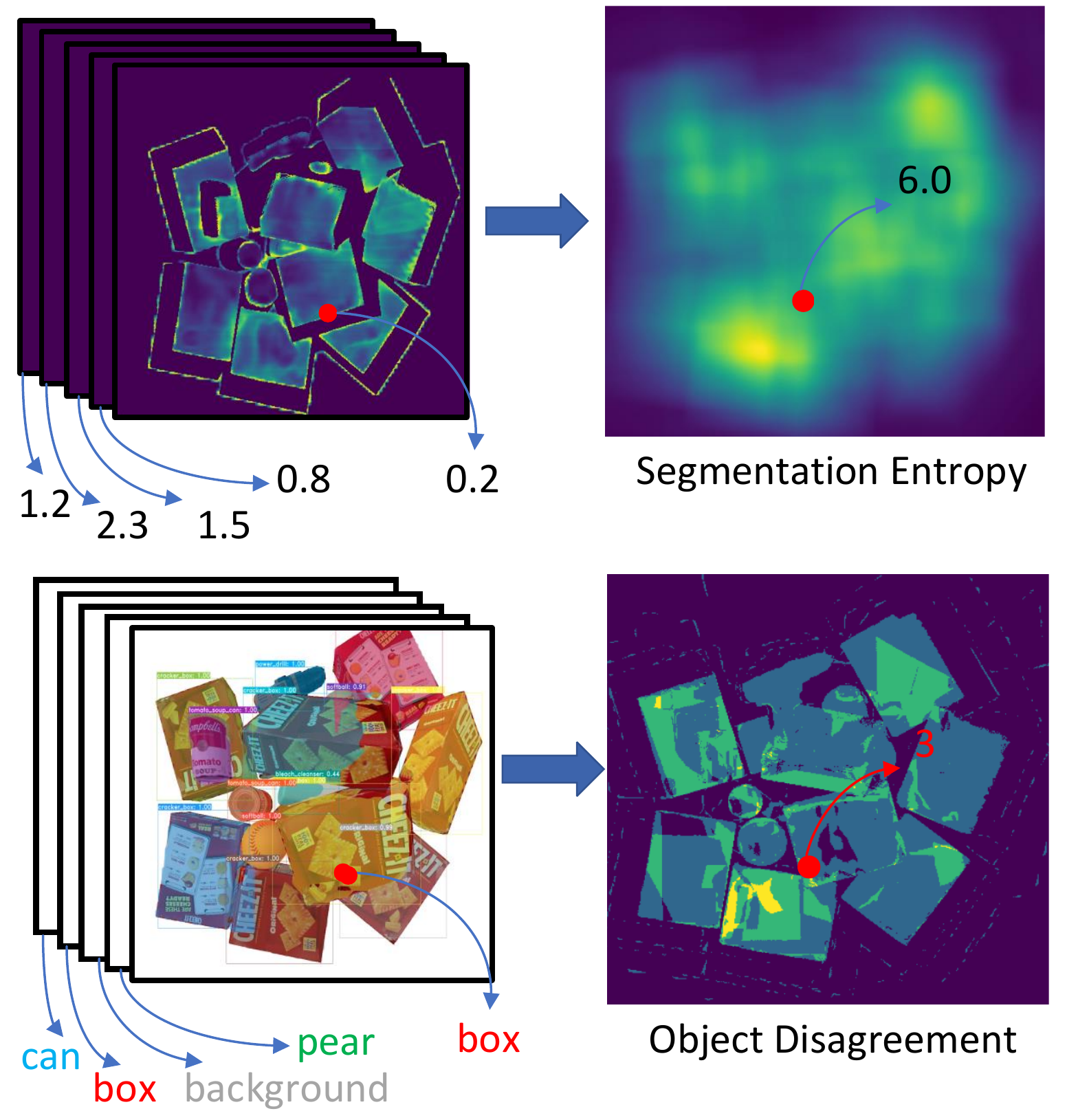}
	\caption{Computation of recognition uncertainty including instance segmentation entropy and multi-view object disagreement based on (\ref{segmentation_entropy}) and (\ref{object_inconsistency}) respectively. For the former component, the instance segmentation entropy in different views where the pixels are projected to the same location in the top-down direction is summed to demonstrate the entropy map. For the latter regard, the predicted foreground labels of pixels that occupy the same location in the top-down direction are counted to reveal the object disagreement that indicates the object occlusion.  }
	\vspace{-0.5cm}    
	\label{uncertainty}
\end{figure}

The recognition uncertainty is employed to generate pushing for interactive exploration, and the pushing action is implemented by moving the closed gripper along with a straight line in parallel with the tabletop. Therefore, the instance segmentation entropy map for RGB images from different views should be projected into the top-down direction for evaluating the recognition uncertainty:
\begin{align}\label{segmentation_entropy}
U_{ij}^{seg} = \sum_{k=1}^{K}\sum_{rs}u_{rs}^k\cdot I(P_{td}(p_{rs}^k)\subset p_{ij}^{td})
\end{align}where $U_{ij}^{seg}$ demonstrates the instance segmentation entropy for the pixel in the $i_{th}$ row and $j_{th}$ column of the top-down view, and $P_{td}(x)$ shows the coordination where the center of the pixel $x$ is mapped to the top-down view. $p_{rs}^k$ and $p_{ij}^{td}$ respectively represent the pixel in the $r_{th}$ row and $s_{th}$ column of the $k_{th}$ view and that in the $i_{th}$ row and $j_{th}$ column observed from the top-down direction. Moreover, $I(x)$ means the indicator function that equals to one for true $x$ and zero otherwise. The instance segmentation entropy contributed by $p_{rs}^k$ is defined in the following:
\begin{align}
	&u_{rs}^k=\sum_{t}u_{rs}^{tk,de}+u_{rs}^{tk,se}\\\notag
	&=-\sum_{t}(\sum_{c=1}^{C}p_{rs,c}^{tk,de}\log p_{rs,c}^{tk,de}-(p_{rs,f}^{tk,se}\log p_{rs,f}^{tk,se}+p_{rs,b}^{k,se}\log p_{rs,b}^{tk,se}))
\end{align}where $u_{rs}^{tk,de}$ and $u_{rs}^{tk,se}$ represent the entropy of object detection and foreground segmentation respectively for the $t_{th}$ bounding box containing $p_{rs}^k$ during instance segmentation. $p_{rs,c}^{tk,de}$ means the probability of the $t_{th}$ bounding box including $p_{rs}^k$ being classified into the $c_{th}$ class, and $p_{rs,f}^{tk,se}$ and $p_{rs,b}^{tk,se}$ respectively stand for the foreground and background probability in the same bounding box for segmentation. The entropy of object detection depicts the uncertainty for assigning the bounding box to a certain class, where high detection entropy indicates ambiguous existence of objects \cite{wang2021learning}. Meanwhile, the entropy of foreground segmentation within the bounding boxes also reveals the ambiguity of the pixel being classified into the detected objects or the background, and high entropy also demonstrates the uncertain object occupancy of pixels. As shown in (\ref{segmentation_entropy}), the entropy contributed by pixels whose centers are mapped inside the same pixel in the top-down view is summed to demonstrate the overall instance segmentation entropy for subsequent pushing generation.

The object disagreement across views also depict the difference of objects that occupy the same pixels in the top-down view due to the occlusion. Therefore, pixels observed in the top-down direction with high multi-view object disagreement are regarded to have high recognition uncertainty due to the invisibility of occluded objects. The object disagreement is defined by counting the predicted foreground classes of pixels whose center is projected to the same location in the top-down view:
\begin{align}\label{object_inconsistency}
U_{ij}^{obj}=N(\{c_{rs}^k|P_{td}(p_{rs}^k)\subset p_{ij}^{td}\})
\end{align}where $U_{ij}^{obj}$ represents the multi-view object disagreement of the top-down view pixel in the $i_{th}$ row and $j_{th}$ column, and $c_{rs}^k$ means the predicted category of the pixel in the $r_{th}$ row and $s_{th}$ column from the $k_{th}$ view. Meanwhile, $N({s})$ stands for the number of non-repeating elements in the set $\{s\}$. By counting the predicted labels for pixels across views that are projected to the same location in the top-down perspective, the object disagreement that reveals the occlusion in the clutter contributes to the overall recognition uncertainty. 
Fig. \ref{uncertainty} shows an example of computing the uncertainty composed of instance segmentation entropy and multi-view object disagreement. Finally, the recognition uncertainty can be written as follows:
\begin{align}\label{recognition_uncertainty}
	U_{ij} = U_{ij}^{seg} + \lambda U_{ij}^{obj}
\end{align}Similarly, $U_{ij}$ stands for the recognition uncertainty of the pixels in the $i_{th}$ row and $j_{th}$ column for the top-down view, and $\lambda$ represents the hyperparameter that controls the importance of multi-view object disagreement in the overall recognition uncertainty.

\subsection{Pushing Generation for Interative Exploration}
Interactively exploring the cluttered scene is crucial for recognizing all objects in dense clutter since the severe occlusion and high crowdedness that degrades the informativeness of visual clues are alleviated. Meanwhile, the interactive exploration should be efficient to be equipped in a wide variety of downstream tasks such as packing and rearranging without sizable latency. Therefore, we leverage the efficient pushing action for interactive exploration to achieve the optimal accuracy-efficiency trade-off, which is implemented by moving the closed gripper along a straight line in parallel to the tabletop.

The location with high recognition uncertainty requires to be pushed so that informative visual clues are uncovered for accurate object recognition in dense clutter with least number of motions. Let us denote $A_{ij}$ as the region in the gripper size with the left-top corner at the $i_{th}$ row and $j_{th}$ column, and the informativeness score for $A_{ij}$ can be described by the average uncertainty of pixels in the region:
\begin{align}\label{informativeness}
	I_{ij}=\frac{1}{N_{ij}}\sum_{p_{rs}\in A_{ij}}U_{rs}
\end{align} where $I_{ij}$ and $N_{ij}$ stand for the informativeness score and number of pixels for $A_{ij}$ respectively. As we expect the clutter structure can be obviously modified by the generated pushing actions to achieve high motion efficiency, the spatial constraint that the start point of pushing should be low is required for informative visual clue discovering. With the height map from the overhead RGB-D camera, we present spatial constraint evaluation metric to assess the validity score in different locations:
\begin{align}\label{validity}
V_{ij}=-\max\limits_{p_{rs}\in A_{ij}}h_{rs}
\end{align}where $V_{ij}$ demonstrates the validity score for the region $A_{ij}$, and $p_{rs}$ and $h_{rs}$ respectively represent the pixel and the corresponding height in the $r_{th}$ row and $s_{th}$ column. The validity depicts the maximum height in the candidate region which is the limitation for gripper decedent. Higher validity score indicates that the pushing can interact with the target region more significantly because of the larger contact area between the gripper and the objects in the target region. Finally, the start point of pushing is generated with the closed gripper in $A_{ij}^{*}$:
\begin{align}\label{push_generation}
A_{ij}^{*} = \arg\max\limits_{A_{ij}}~I_{ij}+\eta V_{ij}
\end{align}where $\eta$ is the hyperparameter that balances the validity and informativeness score in optimal selection for the start point of pushing.

As for the direction for pushing generation, we enumerate the uncertainty map to select a square in gripper size with the highest average uncertainty as the target region for pushing. Leveraging the average uncertainty aims to smooth the noise caused by data collection and pixel projection, which is eliminated by surrounding pixels without corrupting the information carried by the uncertainty map. We utilize the direction from the start point to the target region as the pushing direction, which aims to effectively modify the structure in the region with the highest uncertainty and provides the informative visual clues. A random direction is chosen when the start point and the target region are too close in order to avoid invalid interaction. The pushing distance is set to a constant and scaled twice if the distance between the start points of consecutive pushing is less than a threshold. By iteratively pushing the clutter with interactive exploration, the severe occlusion among objects and high crowdedness is alleviated with the optimal accuracy-efficiency trade-offs for object recognition in dense clutter.

\section{Experiments}
In this section, we conduct extensive experiments in simulated environments (PyBullet \cite{coumanspython}) to evaluate our Smart Explorer. The goal of the experiments is to verify that (1) our object recognition framework for dense clutter can accurately predict the existed classes and the number of objects for each class, (2) the interactive exploration by pushing actions significantly enhance the recognition performance, (3) the generated pushing actions according to recognition uncertainty and spatial constraint outperforms random pushing with respect to accuracy and efficiency.

\begin{figure}[t]
	\centering
	\includegraphics[scale=0.4]{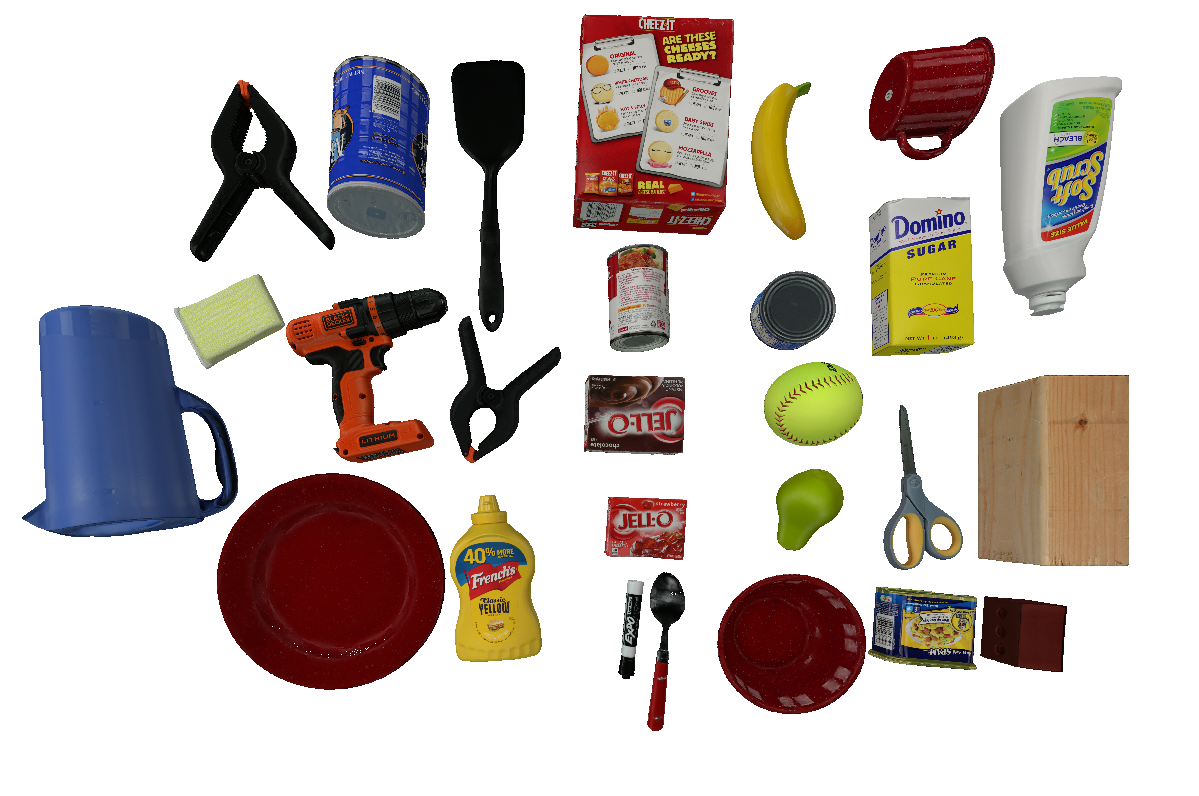}
	\vspace{-0.5cm}
	\caption{The selected subset of objects in our experiments.}
	\label{fig_ycb}
	\vspace{-0.2cm}
\end{figure}

\begin{figure}[t]
	\centering
	\includegraphics[scale=0.32]{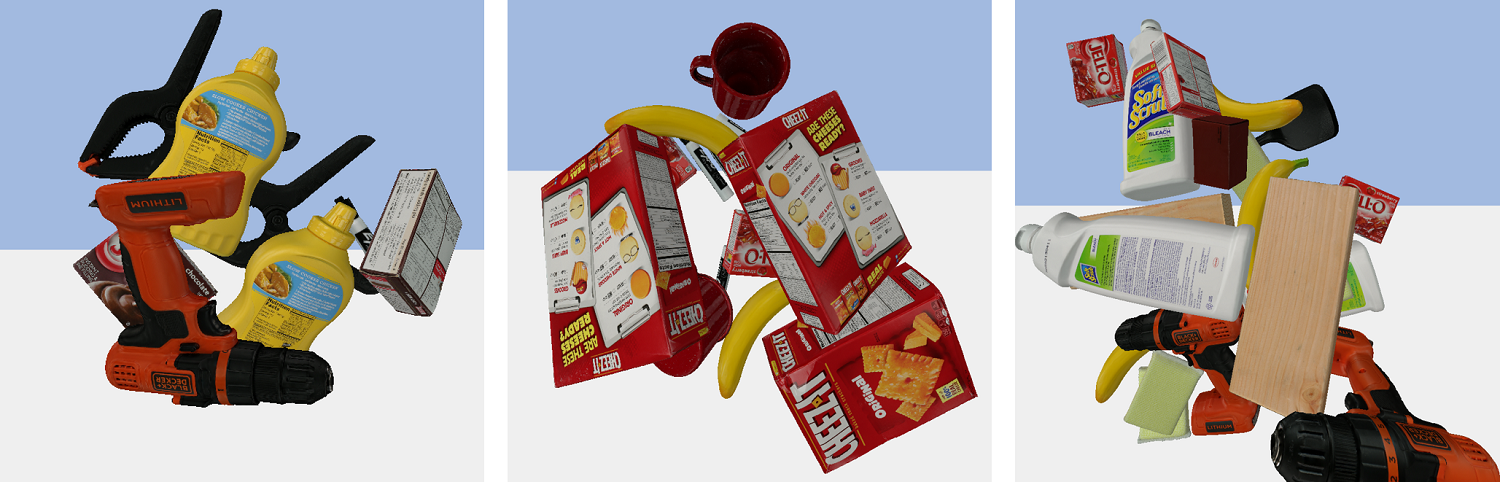}
	\caption{Visualizations of randomly generated scenes, where the easy (left), normal (middle) and hard (right) cases contain 10, 15, and 20 instances.}
	\label{scene_visualization}
	\vspace{-0.5cm}
\end{figure}
%Inspired by previous work, we also divide the test scenarios into random and challenging parts, and all comparative experiments are carried out in fixed cases.

\subsection{Implementation Details} \label{implementation}
The work space is a rectangular area with the size of $1.50m \times 1.00m$ in the simulation environment, and the resolution of each RGB-D vision sensor in the simulated scene is $1440 \times 1024$. In the world coordinate, we set the resolution of each pixel to $2mm$ and the workspace is discretized as a grid of $750 \times 500$ pixels. Except for the overhead camera, the four side-view cameras are evenly distributed in a horizontal plane whose connecting line to the workspace center deviates the vertical direction by $60$ degrees. The size of the region covered by the closed gripper is $2.4^2cm^2$. In order to further reduce the computational cost in pushing generation, the validity and informativeness score shown in (\ref{validity}) and (\ref{informativeness}) are only calculated for start point selection in the square of $0.4^{2}m^{2}$ with the highest average uncertainty.

We utilize the instance segmentation framework Yolact\cite{bolya2019yolact} for RGB image in our Smart Explorer with the segmentation confidence threshold $0.35$, where the IoU threshold to predict a positive bounding box is $0.8$. The Chamfer distance threshold in (\ref{iterative_merge}) that verifies the mergence across different point cloud partitions is set to be $0.001$. The hyperparameters $\lambda$ in (\ref{recognition_uncertainty}) and $\eta$ in (\ref{push_generation}) are determined as $0.1$ and $0.5$. For pushing generation, the distance is set to the constant $0.1$ and scaled twice when the distance between the start points of consecutive pushing is less than the pre-defined threshold $0.05$. The interactive exploration stops until the maximum uncertainty is less than a threshold, and we vary the threshold to acquire different accuracy-efficiency trade-offs. The maximum number of interactive exploration steps is set to $20$ to avoid trivial actions.

The clutter scenes in our experiments are all composed of objects from the YCB dataset \cite{calli2015benchmarking}. The YCB dataset contains $79$ everyday object models with $600$ high-resolution RGB images for each object, and the texture images and point cloud templates for each object is also provided. We select a subset of object classes to construct our scenes to avoid trivial cases according to the following rules: (1) the selected classes have high-quality 3D models in order to prevent sizable noise that significantly harms the segmentation model, (2) the chosen objects should also achieve good visibility in depth map to generate the correct point cloud for object recognition. We finally select $27$ object classes in our experiments with about $20$ instances for each cluttered scene, as shown in Fig. \ref{fig_ycb}. In order to train the instance segmentation model in our Smart Explorer, we collect $1,600$ RGB images from different views and scenes as the dataset. All the experiments are accelerated with one NVIDIA GeForce GTX 3090 GPU.

\begin{table}[t]
	\begin{center}
		\caption{The performance for random clutters (10-20 instances), where $\#$Motions represents the average number of motions.}
		\label{table1}
		\begin{tabular}{m{2cm}<{\centering}|m{1cm}<{\centering}|m{1cm}<{\centering}|m{1cm}<{\centering}|m{1cm}<{\centering}}
			\hline
			Method   & $P(\%)$ & $R(\%)$ & $F_{1}(\%)$ & $\#$Motions  \\
			\hline
			Baseline  & $56.33$ & $74.43$ & $64.13$ & -\\
			\hline
			Random & $58.62$ & $76.38$ & $66.33$ & $5.00$\\
			SCE+Random & $63.13$ & $90.57$ & $74.40$ & $5.00$\\
			Smart Explorer & $77.51$ & $92.23$ & $84.23$ & $5.33$\\
			\hline
				Random & $58.64$ & $76.16$ & $66.26$ & $4.00$\\
			SCE+Random & $60.29$ & $86.23$ & $70.96$ & $4.00$\\
			Smart Explorer  & $74.87$ & $91.23$ & $82.24$ & $4.33$\\
			\hline
				Random & $58.63$ & $76.38$ & $66.34$ & $2.00$\\
			SCE+Random & $59.33$ & $77.43$ & $67.18$ & $2.00$\\
			Smart Explorer  & $65.80$ & $90.56$ & $76.22$ & $1.67$\\
			\hline
		\end{tabular}
	\end{center}
	\vspace{-0.4cm}
\end{table}

\begin{table}[t]
	\begin{center}
		\caption{The performance with different camera settings.}
		\label{camera_setting}
		\begin{tabular}{m{3.2cm}<{\centering}|m{1cm}<{\centering}|m{1cm}<{\centering}|m{1cm}<{\centering}}
			\hline
			Cameras  & $P(\%)$ & $R(\%)$ & $F_{1}(\%)$  \\
			\hline
			1 Overhead & $69.74$ & $72.77$ & $71.22$\\
            \hline
            1 Overhead + 1 Side-view & $71.77$ & $77.94$ & $74.72$\\
            1 Overhead + 2 Side-view & $72.96$ & $77.23$ & $75.03$\\
            1 Overhead + 3 Side-view & $73.89$ & $81.67$ & $77.58$\\
            1 Overhead + 4 Side-view & $75.62$ & $82.23$ & $78.79$\\
			\hline
		\end{tabular}
	\end{center}
	\vspace{-0.8cm}
\end{table}

\begin{figure}[t]
	\centering
	\includegraphics[scale=0.6]{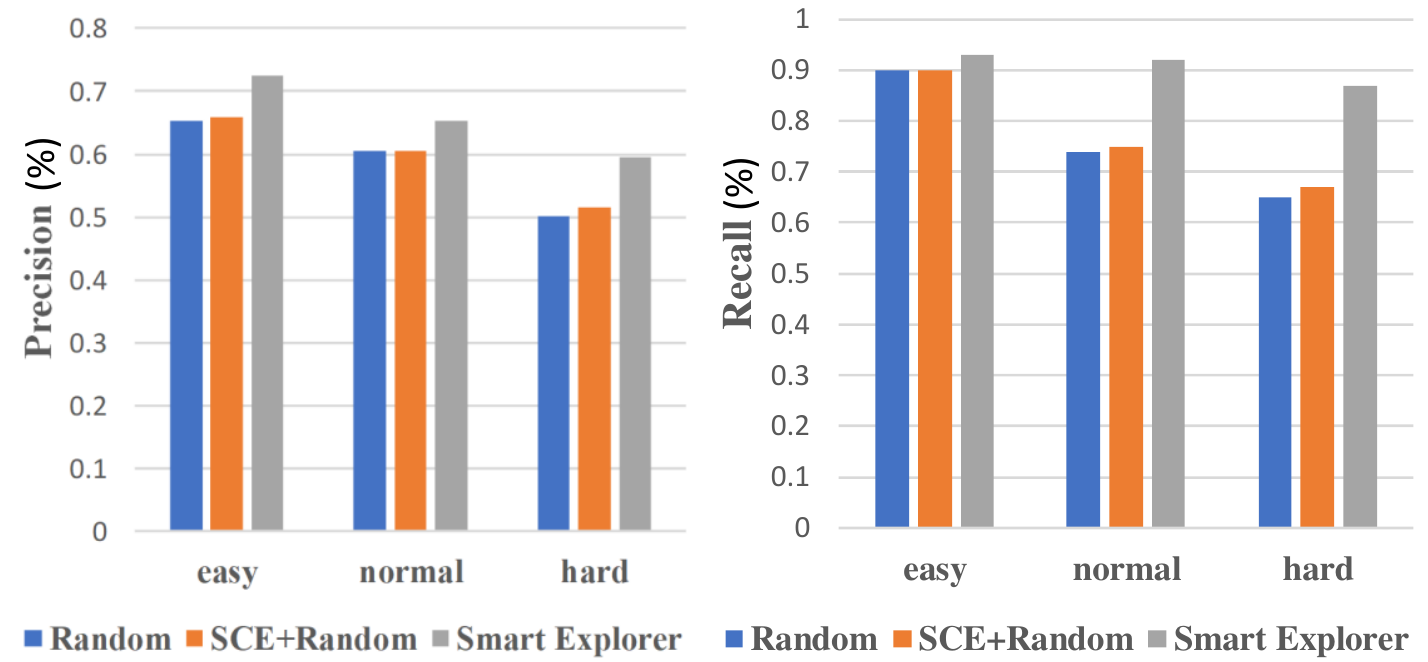}
	\caption{The precision and recall rates for object recognition in dense clutters with different hardness.}
	\vspace{-0.3cm}
	\label{fig}
\end{figure}

\subsection{Evaluation Metrics}
We evaluate our Smart Explorer with respect to the recognition accuracy and motion efficiency. For the first regard, we leverage the precision and recall rate to demonstrate the performance. The number of true positives is defined as the number of predicted objects which is no more than the groundtruth for each class, while the number of false positives mean the extra ones which surpass the groundtruth. Meanwhile, the number of false negatives represent the difference between the groundtruth and predicted numbers for each class. We also report the $F_1$ score  defined as $F_1=2PR/(P+R)$ to further evaluate our Smart Explorer on object recognition, where $P$ and $R$ mean precision and recall rate respectively. For the latter regard, motion efficiency is assessed by the mean number of pushing actions.

\subsection{Results and Discussions}
We execute pushing actions on an UR5 robot equipped with parallel-jaw grippers (robotiq85 gripper), where robot motion planning is implemented by the kinematics module in Pybullet. In order to demonstrate the effectiveness of the interactive exploration in Smart Explorer, we compare our method with the object recognition without pushing and with random pushing. To verify the importance of pushing generation based on recognition uncertainty, we also report the performance with pushing acquired only via the spatial constraint evaluation (SCE). In order to show the performance in different accuracy-efficiency trade-offs, we leveraged three uncertainty thresholds for exploration termination in pushing action generation.

\begin{table}[t]
	\begin{center}
		\caption{The performance for challenging cases (20 instances with severe occlusion).}
		\label{table2}
		\begin{tabular}{m{2cm}<{\centering}|m{1cm}<{\centering}|m{1cm}<{\centering}|m{1cm}<{\centering}|m{1cm}<{\centering}}
			\hline
			Method  & $P(\%)$ & $R(\%)$ & $F_{1}(\%)$ & $\#$Motions  \\
			\hline
			Baseline & $43.09$ & $67.61$ & $52.08$ & -\\
			\hline
			Random & $44.21$ & $68.98$ & $53.90$ & $7.00$\\
			SCE+Random & $49.06$ & $73.39$ & $58.35$  & $7.00$\\
			Smart Explorer & $60.64$ & $86.53$ & $71.19$ & $6.88$\\
			\hline
			Random & $43.84$ & $69.10$ & $53.65$ & $3.00$\\
			SCE+Random & $45.84$ & $70.38$ & $54.99$  & $3.00$\\
			Smart Explorer & $55.28$ & $82.93$ & $66.18$ & $2.87$\\
			\hline
			Random & $43.37$ & $67.83$ & $52.91$ & $2.00$\\
			SCE+Random & $44.18$ & $69.22$ & $53.94$  & $2.00$\\
			Smart Explorer & $47.15$ & $76.25$ & $57.85$ & $1.25$\\
			\hline
		\end{tabular}
	\end{center}
	\vspace{-0.8cm}
\end{table}

\begin{figure}[t]
	\centering
	\includegraphics[scale=0.4]{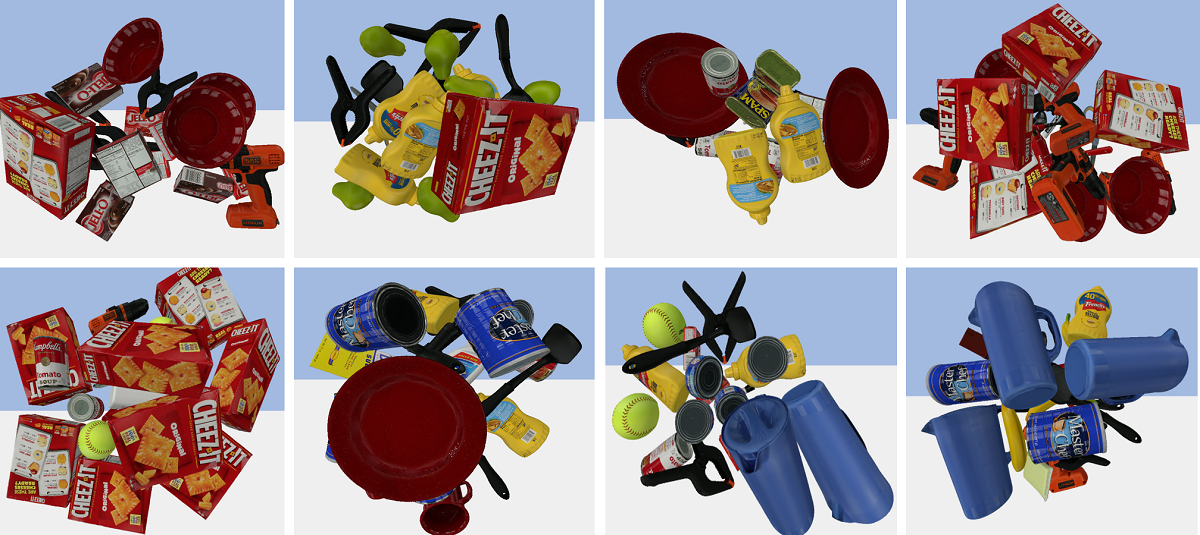}
	\caption{Visualizations of eight challenging cases. Objects in each scene are manually placed to expect a higher degree of dense occlusion, where each case contains about 20 instance objects.}
	\label{fig2}
	\vspace{-0.3cm}
\end{figure}

\subsubsection{Random clutters}
Objects are dropped sequentially into the workspace to generate random clutters, where the landing point of each object is randomly selected. Since the clutter density is positively related to the difficulty of object recognition, we set up the clutter with $10$, $15$ and $20$ instances for easy, normal and hard scenarios for object recognition as shown in Fig. \ref{scene_visualization}. The experimental results in the random clutters are shown in Table \ref{table1}, where our Smart Explorer significantly enhances the precision, recall and $F_1$ score by $9.47\%$ ($65.80\%$ vs. $56.33\%$), $16.13\%$ ($90.56\%$ vs. $74.43\%$) and $12.09\%$ ($76.22\%$ vs. $64.13\%$) compared with the object recognition without interaction by taking only $1.67$ mean actions. Fig. \ref{fig} also shows the precision and recall rates for object recognition in dense clutters with different hardness. Meanwhile, our Smart Explorer outperforms the random push across different action budget by a large margin, which indicates the effectiveness of the pushing generation based on recognition uncertainty and spatial constraint evaluation. The advantage of our method to the generated pushing only with spatial constraint evaluation clearly verify that reducing the recognition uncertainty benefits the object recognition in clutter by discovering informative visual clues. Table \ref{camera_setting} demonstrates the performance with different numbers of side-view RGB-D cameras for visual information collection, where the side-view cameras are evenly distributed in the same plane depicted in Section \ref{implementation}. The number of motions are set to $5$ for each case, and the object disagreement is not considered in uncertainty for the observation setting that is only composed of the overhead camera. The results suggest that our Smart Explore still achieves promising performance in the deployment scenarios where insufficient cameras are provided for visual perception.

\begin{figure}[t]
	\centering
	\includegraphics[scale=0.7]{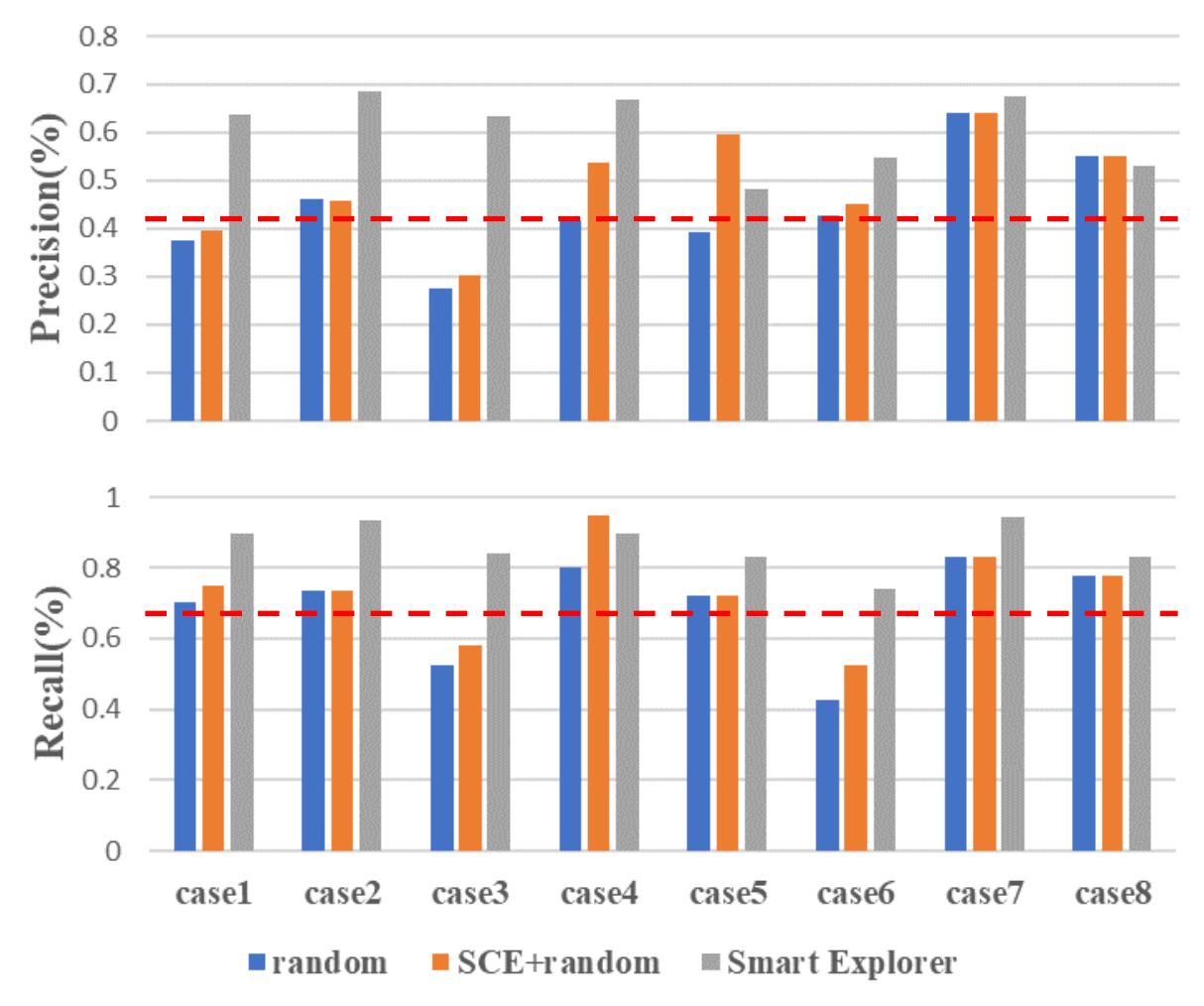}
	\vspace{-0.7cm}
	\caption{The precision and recall rates across eight challenging cases, where the red dashed line represent the average performance of object recognition without interaction.}
	\label{fig4}
    \vspace{-0.6cm}
\end{figure}

\begin{figure}[t]
	\centering
	\includegraphics[scale=0.28]{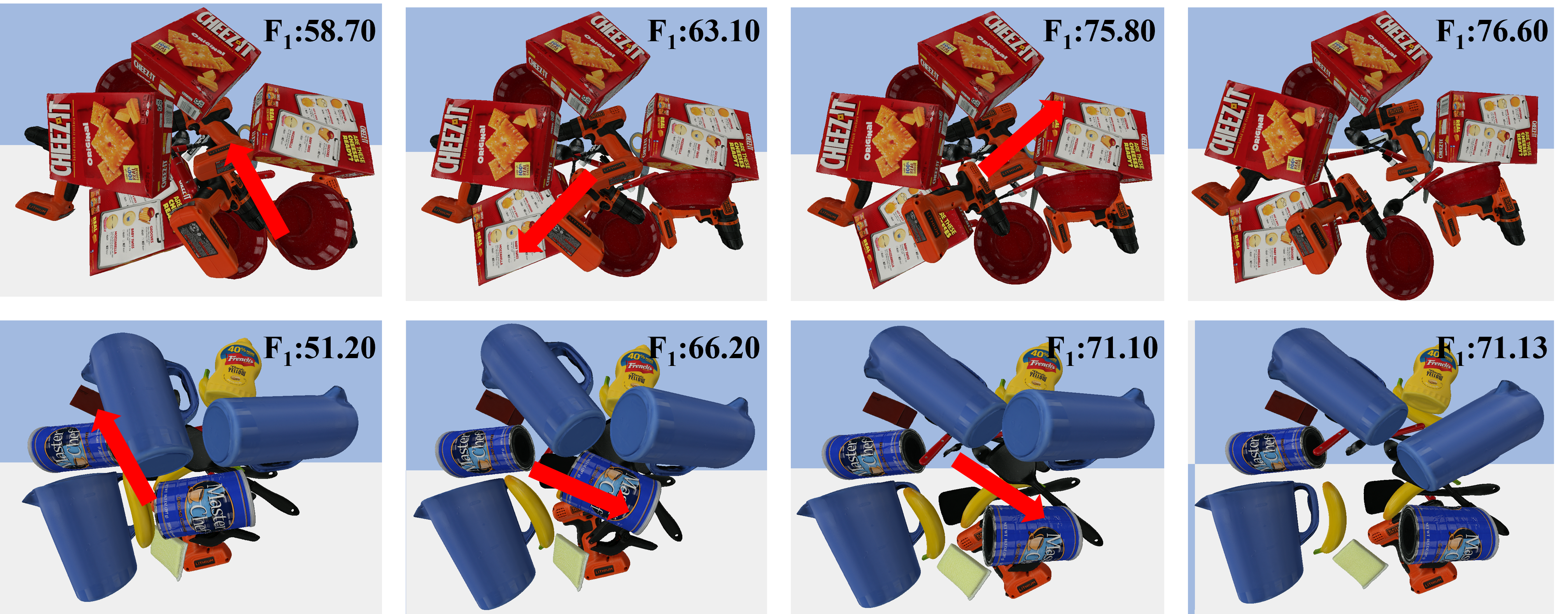}
	\caption{Visualizations of interactive exploration processes, where each row shows a sequence in our Smart Explorer. The red arrow depicts the start point, direction and distance of the generated pushing actions.}
	\label{fig_vg}
	\vspace{-0.5cm}
\end{figure}

\subsubsection{Challenging clutters}
To further verify the effectiveness of our Smart Explorer for object recognition in dense clutter, we evaluate our method in eight challenging scene cases with high object density. The challenging scenarios are constructed manually with dense occlusion among objects, where Fig. \ref{fig2} visualizes the challenging cases respectively. Table \ref{table2} depicts the average performance of different strategies including the precision, recall, $F_1$ score and the mean number of pushing actions in challenging scenarios.
Fig. \ref{fig4} demonstrates the counterparts for each case. For the clutter with extremely high object density, directly recognizing objects acquires much lower accuracy compared with that in random clutters. Randomly generated pushing actions that with spatial constraint evaluation can only strengthen the accuracy by a negligible margin due to the ignorance informativeness in different regions to boost recognition. On the contrary, our Smart Explorer enhances the $F_1$ score by $12.19\%$($55.28\%$ vs. $43.09\%$) with only $2.87$ mean pushing actions, which indicates the practicality in realistic applications with extremely dense clutters. Fig. \ref{fig_vg} visualizes examples of interactive exploration in our Smart Explorer, where dense regions with the highest ambiguity are broken up for informative visual clue discovery.

\section{Conclusion}
In this paper, we have presented Smart Explorer for object recognition in dense clutters, where the existed classes and the number of objects for each class are predicted. We assign the 2D instance segmentation labels to obtain the instance-wise point cloud partition for object recognition, and generate pushing actions for effective interactive exploration to reduce the recognition uncertainty for informative visual clue discovery. Iterative recognition and exploration provides the optimal accuracy-efficiency trade-offs for object recognition in dense clutter. Extensive experiments have demonstrated the effectiveness and efficiency of Smart Explorer.

\section*{Acknowledgements}
This study was supported by the National Natural Science Foundation of China (No. 61976023).

{
	\bibliographystyle{ieee}

}

\end{document}